\documentclass[11pt]{article}
\usepackage{jmlr2e}
\usepackage{amsmath,amsfonts,amscd,amsthm,amssymb}
\usepackage{graphicx,subfigure,epstopdf,float} 
\usepackage{natbib}
\usepackage{algorithm}
\usepackage{algorithmic}
\usepackage[symbol]{footmisc}
\usepackage{url}
\usepackage{hyperref}

\usepackage{array,booktabs,paralist}
\makeatletter
\newenvironment{subtheorem}[1]{%
  \def\subtheoremcounter{#1}%
  \refstepcounter{#1}%
  \protected@edef\theparentnumber{\csname the#1\endcsname}%
  \setcounter{parentnumber}{\value{#1}}%
  \setcounter{#1}{0}%
  \expandafter\def\csname the#1\endcsname{\theparentnumber\alph{#1}}%
  \ignorespaces
}{%
  \setcounter{\subtheoremcounter}{\value{parentnumber}}%
  \ignorespacesafterend
}
\makeatother
\newcounter{parentnumber}

\newtheorem{theorem}{Theorem}

\newtheorem{hypothesis}{Hypothesis}
\newtheorem{mydefinition}{Definition}
\newtheorem{example}{Example}
\newtheorem{assumption}{Assumption}
\newtheorem{lemma}[theorem]{Lemma}

\newtheorem{proposition}[theorem]{Proposition}

\newcommand{\Modified}{Partial}
\newcommand{\modified}{partial}

\renewcommand{\c}{\mathcal}
\renewcommand{\b}{\mathbb}
\newcommand{\cV}{\mathcal{V}}
\newcommand{\cR}{\mathcal{R}}
\newcommand{\cL}{\mathcal{L}}
\newcommand{\cE}{\mathcal{E}}
\newcommand{\cM}{\mathcal{M}}
\newcommand{\cN}{\mathcal{N}}
\newcommand{\cT}{\mathcal{T}}

\newcommand{\cG}{\mathcal{G}}

\newcommand{\bS}{\mathbb{S}}
\newcommand{\bP}{\mathbb{P}}
\newcommand{\bE}{\mathbb{E}}
\newcommand{\bR}{\mathbb{R}}
\newcommand{\bA}{\mathbb{A}}

\newcommand{\Rmn}{\mathbb{R}^{d_1\!\times d_2}}
\newcommand{\Bmn}{\mathbb{B}^{d_1d_2}}
\newcommand{\Smn}{\mathbb{S}^{d_1d_2-1}}
\newcommand{\ksp}{\text{k--sp}}
\newcommand{\ds}{\text{ds}}
\newcommand{\gds}{\text{re}}
\newcommand{\cn}{\text{cn}}
\newcommand{\op}{\text{op}}
\newcommand{\argmin}[1]{\underset{#1}{\text{argmin}}}

\newcommand{\enum}[1]{\begin{asparaenum}#1\end{asparaenum}}
\newcommand{\ibullet}[1]{\begin{asparaitem}#1\end{asparaitem}}
\newcommand{\innerprod}[2]{\langle #1,#2 \rangle}

\newcommand{\mybox}{\hfill\(\Box\)}
\renewcommand{\hat}{\widehat}
\renewcommand{\nu}{\xi}

\newcommand{\pRSC}{\exp\!{(\!{-c_1w_G^2(\c{E}_\cR)})}}

\newcommand{\s}{k}
\newcommand{\w}{\eta}
\newcommand{\p}{\frac{|\Omega|}{d_1d_2}}
\begin{document}
\title{Unified View of Matrix Completion under General Structural Constraints}
 \author{\name{Suriya Gunasekar} {\raggedleft{\email{suriya@utexas.edu}}}
 \AND
 \name{Arindam Banerjee}  {\raggedleft{\email{banerjee@cs.umn.edu}}}
 \AND
 \name{Joydeep Ghosh} {\raggedleft{\email{ghosh@ece.utexas.edu}}}
 }


\maketitle
\begin{abstract}

In this paper, we present a unified analysis of matrix completion under general low-dimensional structural constraints induced by {\em any} norm regularization.
We consider two estimators for the general problem of structured matrix completion, and provide unified upper bounds on the sample complexity and the estimation error. Our analysis relies on results from generic chaining, and we establish two intermediate results of independent interest: (a) in characterizing the size or complexity of low dimensional subsets in high dimensional ambient space, a certain \textit{\modified}~complexity measure encountered in the analysis of matrix completion problems is characterized in terms of a well understood complexity measure of Gaussian widths, and (b) it is shown that a form of restricted strong convexity holds for matrix completion problems under general norm regularization. 
Further, we provide several non-trivial examples of structures included in our framework, notably  the recently proposed spectral $k$-support norm.


%
\end{abstract}

\sloppy

\section{Introduction}\label{sec:intro}
The task of completing the missing entries of a matrix from an incomplete subset of (potentially noisy) entries is encountered in many applications including  recommendation systems, data imputation, covariance matrix estimation, and sensor localization among others. High dimensional estimation problems, where the number of parameters to be estimated is much higher than the number of observations are traditionally ill--posed. However, under low dimensional structural constraints, such problems  have been extensively studied in the recent literature. The special case of matrix completion problems are particularly ill--posed as the observations are both limited (high dimensional), and the measurements are extremely localized, i.e., the observations consist of individual matrix entries. The localized measurement model, in contrast to random Gaussian or sub--Gaussian measurements, poses additional complications in high dimensional estimation.

For well--posed estimation in high dimensional problems, including  matrix completion, it is imperative that low dimensional structural constraints are imposed on the target.
For matrix completion, the special case of low--rank constraint has been widely studied. Several existing work propose tractable estimators with near--optimal recovery guarantees for (approximate) low--rank matrix completion \citep{candes2009exact,candes2010matrix,recht2011simpler, negahban2012restricted,keshavan2010matrix,keshavan2010noise, koltchinskii2011nuclear,davenport2012bit,klopp2014noisy,klopp2015matrix}. A recent work by \citet{gunasekar2014exponential} addresses the extension to structures with decomposable norm regularization. However, the scope of matrix completion extends for low dimensional structures far beyond simple low--rankness or decomposable norm structures.

In this paper, we present a unified statistical analysis of matrix completion under general low dimensional structures that are induced by {\em any} suitable norm regularization. We provide statistical analysis of two generalized matrix completion estimators, the \textit{constrained norm minimizer}, and the \textit{generalized matrix Dantzig selector} (Section~\ref{sec:estimator}). The main results in the paper (Theorem~\ref{thm:main1}--\ref{thm:main2}) provide unified upper bounds on the sample complexity and estimation error of these estimators for matrix completion under \textit{any} norm regularization. Several existing results on matrix completion with low rank or other decomposable structures can be obtained as  special cases of Theorem~\ref{thm:main1}--\ref{thm:main2}.

Our unified analysis of sample complexity is motivated by recent work on high dimensional estimation using global (sub) Gaussian measurements~\citep{chandrasekaran2012convex,amelunxen2013living,tropp2014convex, banerjee2014estimation,vershynin2014estimation,cai2014geometrizing}. A key ingredient in the recovery analysis of high dimensional estimation involves establishing some variation of a certain Restricted Isometry Property (RIP)  \citep{candes2005decoding} of the measurement operator. It has been shown that such properties are satisfied by Gaussian and sub--Gaussian measurement operators with high probability. 
However, as has been noted before by \citet{candes2009exact}, owing to highly localized measurements,  such conditions are not satisfied in the matrix completion problem, and the  existing results based on global (sub) Gaussian measurements are not directly applicable. In fact, one of the questions we address is: given the radically limited measurement model in matrix completion, by how much would the sample complexity of estimation increase beyond the known sample complexity bounds for global (sub) Gaussian measurements? Our results upper bound the sample complexity for matrix completion to within  a $\log d$ factor  over that for estimation under global (sub) Gaussian measurements~\citep{chandrasekaran2012convex,banerjee2014estimation,cai2014geometrizing}. While the result was previously known for low rank matrix completion using nuclear norm minimization \citep{negahban2012restricted,klopp2014noisy}, with a careful use of results from  generic chaining \citep{talagrand2014upper}, we show that the $\log d$ factor suffices for structures induced by {\em any} norm! As a key intermediate result, we show that a useful form of \textit{restricted strong convexity (RSC)} \citep{negahban2009unified} holds for the localized measurements encountered in matrix completion over error sets arising from general norm regularization. The result substantially generalizes existing RSC results for matrix completion under the special cases of nuclear norm and decomposable norm regularization \citep{negahban2012restricted,gunasekar2014exponential}.


For our analysis, we use tools from generic chaining \citep{talagrand2014upper} to characterize the main results (Theorem~\ref{thm:main1}--\ref{thm:main2})  in terms of the \textit{Gaussian width}  (Definition~\ref{def:gwidth}) of certain \textit{error sets}. Gaussian widths provide a powerful geometric characterization for quantifying the complexity  of a structured low dimensional subset in a high dimensional ambient space. Numerous tools have been developed in the literature for bounding the Gaussian width of structured sets.
A unified characterization of results in terms of Gaussian width has the advantage that this literature can be readily leveraged to derive new recovery guarantees for matrix completion under suitable structural
constraints (Appendix $D.2$).

In addition to the theoretical elegance of such a unified framework, identifying useful but potentially non--decomposable low dimensional structures is of significant practical interest. The broad class of structures enforced through symmetric convex bodies and symmetric atomic sets \citep{chandrasekaran2012convex} can be analyzed under this paradigm (Section~\ref{sec:special}). Such specialized structures can  capture the constraints in certain applications better than simple low--rankness. In particular, we discuss in detail, a non--trivial example of the \textit{spectral $k$--support norm} introduced by \citet{mcdonald2014new}.

\noindent  To summarize the key contributions of the paper:
\begin{asparaitem}
\item Theorem~\ref{thm:main1}--\ref{thm:main2} provide unified upper bounds on sample complexity and estimation error for matrix completion estimators using general  norm regularization: a substantial generalization of the existing results on matrix completion under structural constraints. 
\item Theorem~\ref{thm:main1} is applied to derive statistical results for the special case of matrix completion under spectral $k$--support norm regularization.
\item (a) An intermediate result, Theorem~\ref{thm:rsc} shows that under any norm regularization, a variant of Restricted Strong Convexity (RSC) holds in the matrix completion setting with extremely localized measurements. Further, a certain \textit{\modified}~measure of complexity of a set is  encountered in matrix completion analysis \eqref{eq:modified}. (b) Another intermediate result, Theorem~\ref{thm:gaussianWidth} provides bounds on the \textit{\modified}~complexity measures in terms of a better understood complexity measure of Gaussian width. These intermediate results are of independent interest beyond the scope of the paper.
\end{asparaitem}

\textbf{Notations and Preliminaries}

Indexes $i,j$ are typically used to index rows and columns respectively of matrices, and index $\s$ is used to index the observations. $e_i$, $e_j$, $e_\s$, etc.  denote the standard basis in appropriate dimensions\footnote{for brevity we omit the explicit dependence of dimension unless necessary}.
 Notation $G$ and $g$ are used to denote a matrix and vector respectively,  with independent standard Gaussian random variables.  
$\bP(.)$  and $\bE(.)$ denote the probability of an event and the expectation of a random variable, respectively. Given an integer $N$, let $[N]=\{1,2,\ldots, N\}$.
Euclidean norm in a vector space is denoted as $\|x\|_2=\sqrt{\innerprod{x}{x}}$. 
For a matrix $X$ with singular values $\sigma_1\ge\sigma_2\ge\ldots$,  common norms include the \textit{Frobenius norm} $\|X\|_F=\sqrt{\sum_i\sigma_i^2}$, the \textit{nuclear norm} $\|X\|_*=\sum_i\sigma_i$, the \textit{spectral norm} $\|X\|_\op=\sigma_1$, and the \textit{maximum norm} $\|X\|_\infty=\max_{ij}|X_{ij}|$. Also let, $\Smn=\{X\in\Rmn:\|X\|_F=1\}$ and $\Bmn=\{X\in\Rmn:\|X\|_F\le1\}$. 
Finally, given a norm $\c{R}(.)$ defined on a vectorspace $\cV$, its \textit{dual norm} is given by $\c{R}^*(X)=\sup_{\c{R}(Y)\le 1}\langle X,Y\rangle$.


\begin{mydefinition}[Gaussian Width]  \label{def:gwidth}
\normalfont  Gaussian width of a set $S\subset\Rmn$ is a widely studied measure of  complexity of a subset in high dimensional ambient space and is given by:
\begin{equation}w_G(S)=\bE_G\underset{X\in S}{\sup} \innerprod{X}{G},
\label{eq:gwidth}
\end{equation}
where recall that $G$ is a matrix of independent standard Gaussian random variables. Some key results on Gaussian width are discussed in Appendix~$D.2$.
\end{mydefinition}



\begin{mydefinition}[Sub--Gaussian Random Variable \citep{vershynin2010introduction}]\label{def:subGauss}\normalfont The sub--Gaussian norm of a random variable $X$ is given by: $\|X\|_{\Psi_2}=\sup_{p\ge 1}{p^{-1/2}}(\bE|X|^p)^{1/p}$.
$X$ is \textit{$b$--sub--Gaussian} if $\|X\|_{\Psi_2}\le b<\infty$.

Equivalently, $X$ is sub--Gaussian if one of the following conditions are satisfied for some constants  $k_1$, $k_2$, and $k_3$ [Lemma $5.5$ of \citep{vershynin2010introduction}].\\
\begin{inparaenum}[(1)]
\item \; $\forall p\ge 1$,  $(\bE|X|^p)^{1/p}\le b\sqrt{p}$, \quad\quad
\item $\forall t>0$, $\bP(|X|>t)\le e^{1-{{t^2}/{k_1^2b^2}}}$,\\
\item \; $\bE[e^{k_2{X^2}/{b^2}}]\le e$, or\quad\quad \quad\quad\quad\;\,
\item if $\b{E}X=0$, then $\forall s>0$,  $\bE[e^{sX}]\le e^{k_3s^2b^2/2}$.
\end{inparaenum}
\end{mydefinition}

\begin{mydefinition}[Restricted Strong Convexity (RSC)] \label{def:rsc}
\normalfont A function $\mathcal{L}$ is said to satisfy \textit{Restricted Strong Convexity (RSC)}  at $\Theta$ with respect to a subset $S$, if for some \textit{RSC parameter} $\kappa_\mathcal{L}>0$,
\begin{equation}
\forall\Delta\in S, \mathcal{L}(\Theta+\Delta)-\mathcal{L}(\Theta)-\langle\nabla
\mathcal{L}(\Theta),\Delta\rangle\ge\kappa_\mathcal{L}\|\Delta\|_F^2.
\label{eq:RSC}
\end{equation}
\end{mydefinition}



\begin{mydefinition}[Spikiness Ratio \citep{negahban2012restricted}] \normalfont For $X\!\in\!\Rmn,$ a measure of its ``spikiness"  is given by:
\begin{equation}
\alpha_{\text{sp}}(X)=\frac{\sqrt{d_1d_2}\|X\|_{\infty}}{\|X\|_F}.
\label{eq:spiky}
\end{equation}
\end{mydefinition}

\begin{mydefinition}[Norm Compatibility Constant \citep{negahban2009unified}] \label{def:psi}
\normalfont The compatibility constant of a norm $\cR:\mathcal{V}\to\bR$ under a closed convex cone $\mathcal{C}\subset\mathcal{V}$ is defined as follows:
\begin{equation}
\Psi_\cR(\mathcal{C})=\underset{X\in \mathcal{C}\setminus \{0\}}{\sup}\frac{\cR(X)}{\|X\|_F}.
\label{eq:Psi}
\end{equation}
\end{mydefinition} 
\section{Structured Matrix Completion} \label{sec:setup}
 Denote the ground truth target matrix as $\Theta^*\in\Rmn$; let $d\!=\!d_1\!+d_2$. In the noisy matrix completion, observations consists of individual entries of $\Theta^*$ observed through an additive noise channel. \\
\textbf{Sub--Gaussian Noise:} Given, a list of independently sampled standard basis $\Omega=\{ E_\s=e_{i_\s}e_{j_\s}^\top:i_\s\in[d_1],j_\s\in[d_2]\}$ with potential duplicates, 
 observations $(y_\s)_\s\in\bR^{|\Omega|}$ are given by:
\begin{equation}
y_\s=\innerprod{\Theta^*}{E_\s}+\nu \w_\s, \text{ for }\s=1,2,\ldots,|\Omega|,
\label{eq:obs}
\end{equation} 
where $\w\in\bR^{|\Omega|}$ is the noise vector of  independent sub--Gaussian random variables with $\bE[\w_\s]=0$ and $\text{Var}(\w_\s)=1$, and ${\nu}^2$ is scaled variance of noise per observation. Further let $\|\w_{\s}\|_{\Psi_2}\le b$ for a constant $b$ (recall $\|.\|_{\Psi_2}$ from Definition \ref{def:subGauss}). 
Also, without loss of generality, assume normalization $\|\Theta^*\|_F=1$. \\
\textbf{Uniform Sampling:} Assume that the entries in $\Omega$ are drawn independently and uniformly: 
\begin{equation}
	 E_\s\sim\text{uniform}\{e_ie_j^\top: i\in[d_1],j\in[d_2]\}, \text{ for }E_\s\in\Omega.
\label{eq:sampling}
\end{equation}
Let $\{e_k\}$ be the standard basis of $\bR^{|\Omega|}$. 
Given $\Omega$, define $P_\Omega:\Rmn\to\bR^{|\Omega|}$ as:
\begin{equation}
P_\Omega(X)={\textstyle\sum_{\s=1}^{|\Omega|}}\innerprod{X}{E_\s}e_\s
\end{equation}
\textbf{Structural Constraints}
For  matrix completion  with $|\Omega|\!<\! d_1d_2$, low dimensional structural constraints on $\Theta^*$ are necessary for well--posedness. We consider a generalized constraint setting wherein for some low--dimensional  \textit{model space}  $\cM$, $\Theta^*\in\cM$ is enforced through  a surrogate \textit{norm regularizer} $\cR(.)$. We make no further assumptions on $\c{R}$ other than it being a norm in $\Rmn$.

\textbf{Low Spikiness} In matrix completion under uniform sampling model, further restrictions  on $\Theta^*$ (beyond low dimensional structure) are required to ensure that the most informative entries of the matrix are observed with high probability \citep{candes2009exact}. Early work assumed stringent matrix incoherence conditions for low--rank completion to preclude such matrices \citep{candes2010matrix,keshavan2010matrix,keshavan2010noise}, while more recent work~\citep{davenport2012bit,negahban2012restricted}, relax these assumptions to a more intuitive restriction of the {spikiness ratio}, defined in \eqref{eq:spiky}. However, under this relaxation only an approximate recovery is typically guaranteed in low--noise regime, as opposed to near exact recovery under incoherence  assumptions \citep{negahban2012restricted,davenport2012bit}. 
\begin{assumption}[{Spikiness Ratio}]\label{ass:spikiness}
There exists $\alpha^*>0$, such that \\ ${\quad\quad\quad\quad\quad\quad\quad\quad\quad\quad\quad\quad\|\Theta^*\|_{\infty}=\alpha_{\text{\normalfont sp}}(\Theta^*)\frac{\|\Theta^*\|_F}{\sqrt{d_1d_2}}\le\frac{\alpha^*}{\sqrt{d_1d_2}}.}$\mybox
\end{assumption}
\subsection{Special Cases and Applications} \label{sec:special}
We briefly introduce some interesting examples of structural constraints with practical applications.  

\begin{example}[Low Rank and Decomposable Norms]\normalfont 
\textit{Low--rankness} is the most common structure used in many matrix estimation problems including collaborative filtering, PCA, spectral clustering, etc. Convex estimators for low--rank matrix completion using nuclear norm  $\|\Theta\|_*$ regularization has been widely studied statistically \citep{candes2009exact,candes2010matrix,recht2011simpler, negahban2012restricted,keshavan2010matrix,keshavan2010noise, koltchinskii2011nuclear,davenport2012bit,klopp2014noisy,klopp2015matrix}. 
A recent work by \citet{gunasekar2014exponential} extends the   analysis of  matrix completion to general decomposable norms: norms $\cR$, such that $\forall X,Y\!\in\!(\c{M},\c{M}^\perp), \cR(X\!+\!Y)\!=\!\cR(X)\!+\!\cR(Y)$.
\end{example}

\begin{example}[Spectral $\boldsymbol{k}$--support Norm]\normalfont A non--trivial and significant example of norm regularization that is not decomposable is the \textit{spectral $k$--support} norm recently introduced by \citet{mcdonald2014new}. Spectral $k$--support norm is essentially the vector $k$--support norm (overlapping group lasso penalty over all groups for $k$--sparsity) \citep{argyriou2012sparse} applied on the singular values $\sigma(\Theta)$ of a matrix $\Theta\in\Rmn$. Without loss of generality, let $\bar{d}=d_1=d_2$. \\
Let $\cG_k=\{g\subseteq [\bar{d}]: |g|\le k\}$ be the set of all subsets $[\bar{d}]$ of cardinality at most $k$, and let $\cV(\mathcal{G}_k)=\{(v_g)_{g\in\cG_k}: v_g\in\bR^{\bar{d}}, \text{supp}(v_g)\subseteq g\}$. The spectral $k$--support norm is given by:
\begin{equation}
\|\Theta\|_\ksp=\inf_{v\in\cV(\mathcal{G}_k)}\Big\{\sum_{g\in\cG_k}\|v_g\|_2:\sum_{g\in\cG_k}v_g=\sigma(\Theta)\Big\},
\label{eq:ksupp}
\end{equation}
\citet{mcdonald2014new} showed that spectral $k$--support norm is a special case of  \textit{cluster norm} \citep{jacob2009clustered}. It was further shown that in multi--task learning, wherein the tasks (columns of $\Theta^*$) are assumed to be clustered into dense groups,  the cluster norm provides a trade--off between intra--cluster variance, (inverse) inter--cluster variance, and the norm of the task vectors. Both \citet{jacob2009clustered} and \citet{mcdonald2014new} demonstrate superior empirical performance of cluster norms (and $k$--support norm) over traditional trace norm and spectral elastic net minimization on bench marked matrix completion and multi--task learning datasets. 
However, statistical analysis of consistent matrix completion using spectral $k$--support norm regularization has not been previously studied. In Section~\ref{sec:mainKsupport}, we discuss  the consequence of our main theorem for this non--trivial special case.  
\end{example}

\begin{example}[Additive Decomposition]\normalfont Elementwise sparsity is a common structure often assumed in high--dimensional estimation problems. However, in matrix completion, elementwise sparsity conflicts with Assumption~\ref{ass:spikiness} (as well as more traditional incoherence assumptions). Indeed, it is easy to see that with high probability most of the $|\Omega|\ll d_1d_2$ uniformly sampled observations will be zero, and an informed prediction is infeasible. However, elementwise sparse structures can often be modelled within an \textit{additive decomposition} framework, wherein $\Theta^*=\sum_k \Theta^{(k)}$, such that each component matrix $\Theta^{(k)}$ is  in turn structured (e.g. low rank+sparse used for robust PCA \citep{candes2011robust}). In such structures,  there is no scope for recovering sparse components outside the observed indices, and it is  assumed that: $\Theta^{(k)}$ is sparse $\Rightarrow \text{supp}(\Theta^{(k)})\subseteq \Omega$. In such cases, our results are applicable under additional regularity assumptions that enforces non--spikiness on the superposed matrix. A candidate norm regularizer for such structures is the weighted infimum convolution of  individual structure inducing norms \citep{candes2011robust,yang2013dirty}, $$\cR_w(\Theta)=\inf\big\{\sum_kw_k\cR_k(\Theta^{(k)}): \sum_k\Theta^{(k)}=\Theta\big\}.$$
\end{example}

\begin{example}[Other Applications]\normalfont Other potential applications including \textit{cut matrices} \citep{srebro2005rank,chandrasekaran2012convex}, structures induced by \textit{compact convex sets}, norms inducing \textit{structured sparsity assumptions on the spectrum of $\Theta^*$}, etc. can also be handled under the paradigm of this paper. 
\end{example}

\subsection{Structured Matrix Estimator}\label{sec:estimator}
Let $\cR$ be the norm surrogate for the structural constraints on $\Theta^*$, and $\cR^*$ denote its dual norm. 
We propose and analyze two convex estimators for the task of structured matrix completion: \\
\noindent \textbf{Constrained Norm Minimizer}
\begin{equation}
\begin{aligned}
\widehat{\Theta}_{\cn}=&\argmin{{\|\Theta\|_\infty\le\frac{\alpha^*}{\sqrt{d_1d_2}}}}\cR(\Theta)\quad \quad\text{s.t. }\|P_\Omega(\Theta)-y\|_2\le\lambda_{\cn}.
\end{aligned}
\label{eq:estimatorConstrainedNorm}
\end{equation}
\textbf{Generalized Matrix Dantzig Selector}
\begin{equation}
\begin{aligned}
\widehat{\Theta}_{\ds}=&\argmin{{\|\Theta\|_\infty\le\frac{\alpha^*}{\sqrt{d_1d_2}}}}\cR(\Theta)\quad\quad\text{s.t. }\frac{\sqrt{d_1d_2}}{|\Omega|}\cR^*P^*_\Omega(P_\Omega(\Theta)-y)\le\lambda_{\ds},
\end{aligned}
\label{eq:estimatorDantzig}
\end{equation}
where $P_\Omega^*:\bR^{\Omega}\to\Rmn$ is the linear adjoint of $P_\Omega$, i.e. $\innerprod{P_\Omega(X)}{y}=\innerprod{X}{P_\Omega^*(y)}$.\\ 
\textbf{Note:} Theorem~\ref{thm:main1}--\ref{thm:main2} gives consistency results for  \eqref{eq:estimatorConstrainedNorm} and \eqref{eq:estimatorDantzig}, respectively, under certain conditions on the parameters $\lambda_\cn>0$, $\lambda_\ds>0$, and $\alpha^*>1$. In particular, these conditions  assume knowledge of tight bounds on   noise variance $\nu^2$ and spikiness ratio $\alpha_{\text{sp}}(\Theta^*)$. In practice, typically $\nu$ and $\alpha_{\text{sp}}(\Theta^*)$ are  unknown and the parameters are tuned by validating on held out data.
\section{Main Results}\label{sec:main}
We define the following ``restricted" \textit{error cone} and its subset:
\begin{equation}
\cT_\cR=\cT_\cR(\Theta^*)=\text{cone}\{\Delta:\cR(\Theta^*+\Delta)\le \cR(\Theta^*)\},\text{and }\cE_\cR=T_\cR \cap \Smn, 
\label{eq:Tr}
\end{equation}
where recall $\Smn=\{X\in\Rmn:\|X\|_F=1\}$.

Let $\widehat{\Theta}_{\cn}$ and $\widehat{\Theta}_{\ds}$ be the estimates from \eqref{eq:estimatorConstrainedNorm} and \eqref{eq:estimatorDantzig}, respectively.  
If $\lambda_\cn$ and $\lambda_\ds$ are chosen such that ${\Theta}^*$ belongs to the feasible sets in \eqref{eq:estimatorConstrainedNorm} and \eqref{eq:estimatorDantzig}, respectively, then the error matrices $\hat{\Delta}_\cn=\hat{\Theta}_\cn-\Theta^*$ and $\hat{\Delta}_\ds=\widehat{\Theta}_\ds-{\Theta}^*$ are contained in $\cT_\cR$. 

\begin{subtheorem}{theorem} \label{thm:main}
\begin{theorem}[Constrained Norm Minimizer]\label{thm:main1}
Under the problem setup in Section~\ref{sec:setup}, let $\widehat{\Theta}_\cn=\Theta^*+\widehat{\Delta}_\cn$ be the estimate from \eqref{eq:estimatorConstrainedNorm} with $\lambda_\cn=2\nu\sqrt{|\Omega|}$.
For large enough $c_0$,  if $|\Omega|>c_0^2w_G^2(\cE_\cR)\log{d},$
 then there exists an RSC parameter $\kappa_{c_0}>0$ with $\kappa_{c_0}\approx 1-{o}\Big(\frac{1}{\sqrt{\log{d}}}\Big)$, and   constants $c_1$ and $c_2$  such that,  with probability greater than $1\!-\!\pRSC\!-\!2\exp{\!(-c_2w_G^2(\cE_\cR)\log{d})}$,
\[\frac{1}{d_1d_2}\|\widehat{\Delta}_\cn\|_F^2\!\le\! 4\max{\Bigg\{\!\frac{\nu^2}{\kappa_{c_0}},\frac{\alpha^{*2}}{d_1d_2}\sqrt{\frac{c_0^2w_G^2(\cE_\cR)\log{d}}{|\Omega|}\!}\Bigg\}}.
\]
\end{theorem}
\begin{theorem}[Matrix Dantzig Selector] \label{thm:main2}
 Under the problem setup in Section~\ref{sec:setup}, let $\widehat{\Theta}_\ds=\Theta^*+\widehat{\Delta}_\ds$ be the estimate from  \eqref{eq:estimatorDantzig} with $\lambda_\ds\ge2\nu\frac{\sqrt{d_1d_2}}{|\Omega|}\cR^*P_\Omega^*(\w)$. For large enough $c_0$,  if ${|\Omega|>c_0^2w_G^2(\cE_\cR)\log{d},}$
 then there exists an RSC parameter $\kappa_{c_0}>0$ with $\kappa_{c_0}\approx 1-{o}\Big(\frac{1}{\sqrt{\log{d}}}\Big)$, and a constant $c_1$ such that,  with probability greater than $1\!-\!\pRSC$,
\[\frac{1}{d_1d_2}\|\widehat{\Delta}_\ds\|_F^2\!\le\! 16\max{\Bigg\{\!\frac{\lambda_\ds^2\Psi^2_\cR(\cT_\cR)}{\kappa_{c_0}^2},\frac{\alpha^{*2}}{d_1d_2}\sqrt{\frac{c_0^2w_G^2(\cE_\cR)\log{d}}{|\Omega|}\!}\Bigg\}}.\]
\end{theorem}
\normalfont Recall Gaussian width $w_G$ and subspace compatibility constant $\Psi_\cR$ from \eqref{eq:gwidth} and \eqref{eq:Psi}, respectively.
\end{subtheorem}

\noindent \textbf{Remarks:}
\begin{compactenum}[\leftmargin=0pt 1.]
\item If $\cR(\Theta)=\|\Theta\|_*$ and $\text{rank}(\Theta^*)=r$, then $w_G^2(\cE_\cR)\le 3dr$, $\Psi_\cR(\cT_\cR)\le8\sqrt{r}$ and w.h.p ${\frac{\sqrt{d_1d_2}}{|\Omega|}\|P_\Omega^*(\w)\|_2\le2\sqrt{\frac{d\log d}{|\Omega|}}}$  \citep{chandrasekaran2012convex,fazel2001rank,negahban2012restricted}. Using these bounds in Theorem~\ref{thm:main2} recovers near--optimal results for low rank matrix completion under spikiness   \citep{negahban2012restricted}.
\item For both estimators, upper bound on sample complexity is dominated by the square of Gaussian width which is  often considered the \emph{effective dimension} of a subset in high dimensional space  and plays a key role in high dimensional estimation under Gaussian measurement ensembles. The results show that, independent of $\cR(.)$, the upper bound on sample complexity for consistent matrix completion with highly localized measurements is within a $\log{d}$ factor of the known sample complexity of $\sim w_G^2(\cE_\cR)$   for estimation from Gaussian  measurements \citep{banerjee2014estimation,chandrasekaran2012convex, vershynin2014estimation,cai2014geometrizing}.
\item First term in estimation error bounds in Theorem~\ref{thm:main1}--\ref{thm:main2} scales with $\nu^2$ which is the per observation noise variance. The second term is an upper bound on error that arises due to unidentifiability of $\Theta^*$ within a certain radius under the spikiness constraints \citep{negahban2012restricted}; in contrast \citet{candes2010matrix} show exact recovery when $\nu=0$ using more  stringent matrix incoherence conditions.
\item Bound on $\hat{\Delta}_\cn$ from Theorem~\ref{thm:main1} is comparable to the result by \citet{candes2010matrix} for low rank matrix completion under non--low--noise regime, where the first term dominates, and those of  \citep{chandrasekaran2012convex,tropp2014convex} for high dimensional estimation under Gaussian measurements. With a bound on  $w_G^2(\cE_\cR)$, it is easy to specialize this result for new structural constraints. However, this bound is potentially loose and asymptotically converges to a constant error proportional to the noise variance $\nu^2$.
\item The estimation error bound in Theorem~\ref{thm:main2} is typically sharper than that in Theorem~\ref{thm:main1}. However, for specific structures, using application of Theorem~\ref{thm:main2} requires additional bounds on $\cR^*P_\Omega^*(\eta)$ and $\Psi_\cR(\cT_\cR)$ besides $w_G^2(\cE_\cR)$. 
\end{compactenum}
\subsection{\Modified~Complexity Measures}
Recall that $w_G(S)=\bE\sup_{X\in S}\innerprod{X}{G}$. $G\in\Rmn$, and $g\in\b{R}^{|\Omega|}$ denotes a random matrix and vector respectively with each entry sampled independently from standard normal distribution. 
\begin{mydefinition}[\Modified~Complexity Measures]  Given a randomly sampled  $\Omega=\{E_\s\in\Rmn\}$, and a centered random vector $\w\in \bR^{|\Omega|}$, the \textit{\modified~$\w$--complexity measure} of $S$ is given by:
\begin{equation}
w_{\Omega,\w}(S)=\bE_{\Omega,\w}\sup_{X\in S-S}\innerprod{X}{P_\Omega^*(\w)}.
\label{eq:modified}
\end{equation}
\end{mydefinition}
Special cases of $\w$ being a vector of standard Gaussian $g$, or standard Rademacher $\epsilon$ (i.e. $\epsilon_\s\in\{-1,1\}$ w.p. $1/2$) variables, are of particular interest. 

\textbf{Note:} In the case of symmetric $\eta$, like $g$ and $\epsilon$, $w_{\Omega,\w}(S)=2\bE_{\Omega,\w}\sup_{X\in S}\innerprod{X}{P_\Omega^*(\w)}$, and the later expression will be used interchangeably ignoring the constant term. \mybox

\begin{theorem}[\Modified~Gaussian Complexity] \label{thm:gaussianWidth}  Let $S\subseteq\Bmn$ with non--empty interior, and  let $\Omega$ be sampled according to \eqref{eq:sampling}.  
$\exists$ universal constants $k_1$, $k_2$, $K_1$ and $K_2$ such that:
\begin{flalign}
\begin{split}
w_{\Omega,g}({S})&\le k_1\sqrt{\frac{|\Omega|}{d_1d_2}}w_G({S})+k_2\sqrt{\bE_\Omega\sup_{X,Y\in S}\|P_\Omega(X-Y)\|_2^2} \\
w_{\Omega,g}({S})&\le K_1\sqrt{\frac{|\Omega|}{d_1d_2}}w_G({S})+K_2\sup_{X,Y\in S}\|X-Y\|_\infty.
\end{split}
\label{eq:thm2}
\end{flalign}
Also, for centered i.i.d. sub--Gaussian vector $\eta\in\bR^{|\Omega|}$, $\exists $ constant $K_3$ s.t.  $w_{\Omega,\eta}({S})\le K_3w_{\Omega,g}({S})$.
\end{theorem}
\textbf{Note: } For $\Omega\subsetneq [d_1]\times [d_2]$, the second term in \eqref{eq:thm2} is a consequence of the localized measurements. 

\subsection{Spectral $k$--Support Norm}\label{sec:mainKsupport}\label{sec:mainSC}
We introduced spectral $k$--support norm in Section~\ref{sec:special}. The estimators from \eqref{eq:estimatorConstrainedNorm} and \eqref{eq:estimatorDantzig} for spectral $k$--support norm can be efficiently solved via proximal methods using the proximal operators derived in \citet{mcdonald2014new}. We are interested in the statistical guarantees for matrix completion using spectral $k$--support norm regularization. We extend the analysis for upper bounding the Gaussian width of the descent cone for the vector $k$--support norm by \citet{richard2014tight} to the case of spectral $k$--support norm.
 WLOG let $d_1=d_2=\bar{d}$. Let $\sigma^*\in \bR^{\bar{d}}$  be the vector of singular values of $\Theta^*$ sorted in non--ascending order. Let $r\in\{0,1,2,\ldots,k-1\}$ be the unique integer satisfying: $ \sigma^*_{k-r-1}>\frac{1}{r+1}\sum_{i=k-r}^{p}\sigma^*_i\ge\sigma^*_{k-r}.$
Denote $I_2=\{1,2,\ldots,k-r-1\}$ and $I_1=\{k-r,k-r+1,\ldots,s\}$. Finally, for $I\subseteq[\bar{d}]$, 
 $(\sigma^*_I)_i=0 \;\forall i\in I^c$, and $(\sigma^*_I)_i=\sigma^*_i \;\forall i\in I$.
\begin{lemma} \label{lem:ksupp1} If rank of $\Theta^*$ is $s$ and
$\cE_\cR$ is the error set for $\cR(\Theta)=\|\Theta\|_\ksp$, then
\[w_G^2(\cE_\cR)\le s(2\bar{d}-s)+\Big(\frac{(r+1)^2\|\sigma^*_{I_2}\|_2^2}{\|\sigma^*_{I_1}\|_1^2}+|I_1|\Big)(2\bar{d}-s).\]
\end{lemma}
Proof of the above lemma is provided in the appendix. Lemma~\ref{lem:ksupp1}  can be combined with Theorem~\ref{thm:main1}
 to obtain  recovery guarantees for  completion under spectral $k$--support norm.
\section{Discussions and Related Work}\label{sec:discussion} 
\textbf{Sample Complexity:} 
For consistent recovery in high dimensional convex estimation, it is desirable that the descent cone at the target parameter $\Theta^*$ is ``small" relative to the feasible set (enforced by the observations) of the estimator. Thus, it is not surprising that  the sample complexity and estimation error bounds of an estimator depends on a measure of complexity/size of the error cone at $\Theta^*$. Results in this paper are largely characterized in terms of a widely used complexity measure of Gaussian width $w_G(.)$, and can be compared with the literature on  estimation from Gaussian measurements. 

\textbf{Error Bounds:} Theorem~\ref{thm:main1} provides estimation error bounds that depends only on the Gaussian width of the descent cone. In non--low--noise regime, this result is comparable to analogous results of constrained norm minimization  \citep{candes2011robust,chandrasekaran2012convex,tropp2014convex}.  
However, this bound is potentially loose owing to mismatched data--fit term using squared loss, and asymptotically converges to a constant error proportional to the noise variance $\nu^2$.\\
A tighter analysis on the estimation error can be obtained for the matrix Dantzig selector \eqref{eq:estimatorDantzig} from Theorem~\ref{thm:main2}. However, application of Theorem~\ref{thm:main2} requires computing high probability upper bound on $\cR^*P_\Omega^*(\w)$. The literature on norms of random matrices  \citep{edelman1988eigenvalues,litvak2005smallest,vershynin2010introduction, tropp2012user} can be exploited in computing such bounds. Beside, in special cases: if $\cR(.)\ge K\|.\|_*$, then $K\cR^*(.)\le\|.\|_\text{op}$ can be used to obtain asymptotically consistent results.

Finally, under  near zero--noise, the second term in the results of Theorem~\ref{thm:main} dominates, and  bounds  are weaker than that of \citet{candes2011robust,keshavan2010noise} owing to the relaxation of stronger incoherence assumption. 

\textbf{Related Work and Future Directions:} The closest related work is the result on consistency of matrix completion under  decomposable norm regularization by \citet{gunasekar2014exponential}. Results in this paper are a strict generalization to general norm regularized (not necessarily decomposable) matrix completion. We provide non--trivial examples of application where structures enforced by such non--decomposable norms are of interest. 
Further, in contrast to our results that are based on Gaussian width, the RSC parameter in \citet{gunasekar2014exponential} depends on a modified complexity measure $\kappa_\cR(d,|\Omega|)$ (see definition in \citet{gunasekar2014exponential}). An advantage of results based on Gaussian width is that, application of Theorem~\ref{thm:main}  for special cases can greatly benefit from the  numerous tools in the literature for the computation of $w_G(.)$.\\ 
Another closely related line of work is the non--asymptotic analysis of high dimensional estimation under random Gaussian or sub--Gaussian measurements~\citep{chandrasekaran2012convex,amelunxen2013living,tropp2014convex, banerjee2014estimation,vershynin2014estimation,cai2014geometrizing}. However, the analysis from this literature rely on variants of RIP of the measurement ensemble \citep{candes2005decoding}, which is  not satisfied by the the extremely localized measurements encountered in matrix completion \citep{candes2009exact}. In an intermediate result, we establish a form of RSC for matrix completion under general norm regularization: a result that was previously known only for nuclear norm and decomposable norm regularization. 

In future work, it is of interest to derive matching lower bounds on estimation error for matrix completion under general low dimensional structures, along the lines of \citet{koltchinskii2011nuclear} and explore special case applications of the results in the paper. We also plan to derive explicit characterization of $\lambda_\ds$ in terms of Gaussian width of unit balls by exploiting generic chaining results for general Banach spaces \citep{talagrand2014upper}. 
\section{Proof Sketch}\label{sec:proof}
Proofs of the lemmas are provided in the appendix. 
\subsection{Proof of Theorem~\ref{thm:main}}
Define the following set of $\beta$--\textit{non--spiky} matrices in $\Rmn$ for constant $c_0$ from Theorem~\ref{thm:main}:
\begin{equation}
\bA(\beta)\!=\!\Bigg\{X\!:\alpha_\text{sp}(X)=\frac{\sqrt{d_1d_2}\|X\|_\infty}{\|X\|_F}< \beta\Bigg\}.
\label{eq:a}
\end{equation}
\begin{flalign}
&\text{Define,} \quad\quad\quad\quad\quad\quad\quad\quad\beta^2_{c_0}=\sqrt{\frac{|\Omega|}{c_0^2w_G^2(\cE_\cR)\log{d}}}&
\label{eq:beta}
\end{flalign} 
\textbf{Case $\boldsymbol{1}$: Spiky Error Matrix} 
When the error matrix from \eqref{eq:estimatorConstrainedNorm} or \eqref{eq:estimatorDantzig}  has large spikiness ratio,  following bound on  error is immediate  using $\|\widehat{\Delta}\|_\infty\!\le\!\|\widehat{\Theta}\|_\infty\!+\!\|\Theta^*\|_\infty\!\!\le\!\!{2\alpha^*}/{\sqrt{d_1d_2}}$ in \eqref{eq:spiky}. 
\begin{proposition}[Spiky Error Matrix] \label{prop:spiky} For the constant $c_0$ in Theorem~\ref{thm:main1}, if $\alpha_\text{sp}(\widehat{\Delta}_\cn)\notin \bA(\beta_{c_0})$, then $\|\widehat{\Delta}_\cn\|_F^2\le \frac{4\alpha^{*2}}{\beta_{c_0}^2}=4\alpha^{*2}\sqrt{\frac{c_0^2w_G^2(\cE_\cR)\log{d}}{|\Omega|}}$. 
An analogous result also holds for $\hat{\Delta}_\ds$. \mybox
\end{proposition}
\noindent \textbf{Case $\boldsymbol 2$: Non--Spiky Error Matrix} Let 
$\hat{\Delta}_\ds,\hat{\Delta}_\cn\in\bA(\beta_{c_0})$.
Recall from \eqref{eq:obs}, that $y-P_\Omega(\Theta^*)=\nu \w$, where $\w\in\bR^{|\Omega|}$ consists of independent sub--Gaussian random variables with $\bE[\w_\s]=0$,  $\text{Var}(\w_\s)=1$. Further, as $\eta$ is sub--Gaussian, let $\|\w_\s\|_{\Psi_2}\le b$ for a constant $b$.

\subsubsection{Restricted Strong Convexity (RSC)}
Recall $\cT_\cR$ and $\cE_\cR$ from \eqref{eq:Tr}. 
An important step in the proof of Theorem~\ref{thm:main} involves showing that over a useful subset of $\cT_\cR$, a form of RSC \eqref{eq:RSC}  is satisfied by a squared loss penalty. 
\begin{theorem}[Restricted Strong Convexity] \label{thm:rsc} Let $|\Omega|>c_0^2w_G^2(\cE_\cR)\log{d}$,  for large enough constant $c_0$. 
There  exists a RSC parameter $\kappa_{c_0}>0$ with $\kappa_{c_0}\approx 1-{o}\Big(\frac{1}{\sqrt{\log{d}}}\Big)$, and a constant $c_1$ such that, the following holds  w.p.  greater that $1-\pRSC$, 
\[\forall X\in \cT_\cR\cap\bA(\beta_{c_0}), \quad \frac{d_1d_2}{|\Omega|}\|P_\Omega(X)\|_2^2\ge \kappa_{c_0}\|X\|_F^2.
\]
\noindent\normalfont 
Proof in appendix combines  empirical process tools along  with Theorem~\ref{thm:gaussianWidth}.\mybox
\end{theorem}
\subsubsection{Constrained Norm Minimizer}
 \begin{lemma} \label{lem:lambdacn} Under the conditions of Theorem~\ref{thm:main}, let $b$ be a constant such that $\forall \s$,  $\|\w_\s\|_{\Psi_2}\le b$. There exists a universal constant $c_2$ such that, if  $\lambda_\cn\!\ge\!2\nu\sqrt{|\Omega|}$, then w.p. greater than $1-2\exp{(-c_2|\Omega|)}$, $(a)$ $\widehat{\Delta}_\ds\in\cT_\cR$, and $(b)$ ${\|P_\Omega(\hat{\Delta}_\cn)\|_2\!\le\!2\lambda_\cn}$.  \mybox\end{lemma}
Using $\lambda_\cn\!=\!2\nu\sqrt{|\Omega|}$ in \eqref{eq:estimatorConstrainedNorm}, if  $\hat{\Delta}_\cn\!\in\!\mathbb{A}(\beta_{c_0})$, then using Theorem \ref{thm:rsc} and Lemma~\ref{lem:lambdacn}, w.h.p.
\begin{equation}
\frac{\|\hat{\Delta}_\cn\|_F^2}{d_1d_2}\!\le\! \frac{1}{\kappa_{c_0}}\!\frac{\|P_\Omega(\widehat{\Delta}_\cn)\|_2^2}{|\Omega|}{\le}\frac{4\nu^2}{\kappa_{c_0}}.
\label{eq:case2b} 
\end{equation} 
\subsubsection{Matrix Dantzig Selector}
\begin{proposition} \label{lem:lambdads} $\lambda_{ds}\!\ge\! \nu\frac{\sqrt{d_1d_2}}{{|\Omega|}}\cR^*P_\Omega^*(\w)\Rightarrow$ w.h.p. (a) $\widehat{\Delta}_\ds\!\in\!\cT_\cR$; (b) $\frac{\sqrt{d_1d_2}}{|\Omega|}\cR^*P_\Omega^*(P_\Omega(\hat{\Delta}_\ds))\!\le\!2\lambda_\ds$.
\end{proposition} 
Above result follows from optimality of $\hat{\Theta}_\ds$ and triangle inequality. 
Also,
\begin{equation*}
\begin{split}
\frac{\sqrt{d_1d_2}}{|\Omega|}\|P_\Omega(\hat{\Delta}_\ds)\|_2^2\le\frac{\sqrt{d_1d_2}}{|\Omega|}\cR^*P_\Omega^*(P_\Omega(\hat{\Delta}_\ds))\cR(\hat{\Delta}_\ds)\le 
2\lambda_\ds\Psi_\cR(\cT_\cR)\|\hat{\Delta}_\ds\|_F,
\end{split}
\end{equation*}
where recall norm compatibility constant $\Psi_\c{R}(\c{T}_\c{R})$ from  \eqref{eq:Psi}. Finally, using Theorem~\ref{thm:rsc}, w.h.p.
\begin{equation}
\frac{\|\hat{\Delta}_\ds\|_F^2}{d_1d_2}\le\frac{1}{|\Omega|}\frac{\|P_\Omega(\widehat{\Delta}_\ds)\|_2^2}{\kappa_{c_0}}{\le}\frac{4\lambda_\ds\Psi_\cR(\cT_\cR)}{\kappa_{c_0}}\frac{\|\widehat{\Delta}_\ds\|_F}{\sqrt{d_1d_2}}. 
\label{eq:case2} 
\end{equation} 
\subsection{Proof of Theorem~\ref{thm:gaussianWidth}}  Let the entries of $\Omega=\{E_\s=e_{i_\s}e_{j_\s}^\top:\s=1,2,\ldots,|\Omega|\}$ be sampled as in \eqref{eq:sampling}. Recall that $g\in\bR^{|\Omega|}$ is a standard normal vector. 
Define the following random process:
\begin{flalign}
(\mathcal{X}_{\Omega,g}(X))_{X\in {S}}, \text{where }\mathcal{X}_{\Omega,g}(X)=\langle X,P_\Omega^*(g)\rangle={\textstyle \sum_{\s}} \langle X,E_\s\rangle g_\s. &\label{eq:rProc}
\end{flalign}
We start with a key lemma in the proof of Theorem~\ref{thm:gaussianWidth}. Proof of this lemma, provided in Appendix~$B$, uses tools from the broad topic of generic chaining developed in recent works \cite{talagrand1996majorizing,talagrand2014upper}.
\begin{lemma}\label{lem:gammaResult}
For a compact subset $S\subseteq\Rmn$ with non--empty interior, $\exists$ constants $k_1$, $k_2$ such that: \\
{$\displaystyle
\nonumber w_{\Omega,g}({S})=\bE\sup_{X\in {S}}\mathcal{X}_{\Omega,g}(X)\le  k_1\sqrt{\frac{|\Omega|}{d_1d_2}}w_G(S)+k_2\sqrt{\bE\sup_{X,Y\in S}\|P_\Omega(X-Y)\|_2^2}.
$\mybox}

\end{lemma}
\begin{lemma} 
There exists constants $k_3$, $k_4$, such that for compact $S\subseteq\Bmn$ with non--empty interior
$$\bE\sup_{X,Y\in S}\|P_\Omega(X-Y)\|_2^2\le k_3\p w_G^2(S)+k_4(\sup_{X,Y\in S}\|X-Y\|_\infty )w_{\Omega,g}(S)$$
\label{prop:temp}
\end{lemma}
Theorem~\ref{thm:gaussianWidth} follows by combining Lemma~\ref{lem:gammaResult} and Lemma~\ref{prop:temp}, and simple algebraic manipulations using $\sqrt{ab}\le a/2+b/2$ and triangle inequality (See Appendix~$B.4$). 

The statement in Theorem~\ref{thm:gaussianWidth} about partial sub--Gaussian complexity follows from a standard result in empirical process given in Lemma~$11$ in the appendix. \mybox

\textbf{Acknowledgments} We thank the anonymous reviewers for helpful comments and suggestions. S. Gunasekar and J. Ghosh acknowledge funding from NSF grants IIS-1421729, IIS-1417697, and IIS–1116656.
A. Banerjee acknowledges NSF grants  IIS-1447566, IIS-1422557,
CCF-1451986, CNS-1314560, IIS-0953274, IIS-1029711, and NASA grant NNX12AQ39A.

{\small 
\bibliographystyle{plain}
\bibliography{bibliography,latestbib}}
\clearpage
\appendix

\section*{\centering Supplementary Material: Unified View of Matrix Completion under General Structural Constraints}
\begin{center}
{\centering \large Suriya Gunasekar, Arindam Banerjee, Joydeep Ghosh}
\end{center}

\noindent {Note:} Background and preliminaries are provided in Appendix~\ref{sec:prelims}. 
\section{Appendix to Proof of Theorem~\ref{thm:main}}\label{app:rsc}

\subsection{Proof of Theorem \ref{thm:rsc}}
\textbf{Statement of Theorem~\ref{thm:rsc}}:\\
Let $|\Omega|>c_0^2w_G^2(\cE_\cR)\log{d}$,  for large enough constant $c_0$. 
There  exists a RSC parameter $\kappa_{c_0}>0$ with $\kappa_{c_0}\approx 1-{o}\Big(\frac{1}{\sqrt{\log{d}}}\Big)$, and a constant $c_1$ such that, the following holds  w.p.  greater that $1-\pRSC$,
\[\forall X\in \cT_\cR\cap\bA(\beta_{c_0}), \quad \frac{d_1d_2}{|\Omega|}\|P_\Omega(X)\|_2^2\ge \kappa_{c_0}\|X\|_F^2.
\]
\textbf{Proof: }
Recall that $\cT_\cR=\{\Delta:\cR(\Theta^*+\Delta)\le \cR(\Theta^*)$ and $\cE_\cR=\cT_\cR\cap\Smn$. Using the properties of norms, it can be easily verified that for the non--trivial case of $\Theta^*\neq0$, $\cT_\cR$ is a cone with non--empty interior. 

We use Theorem~\ref{thm:gaussianWidth} as a key result  in this proof. \\
Define $\bar{\cE}_\cR=\cT_\cR\cap\Bmn$.\\
$\bar{\c{E}}_\cR\supset {\c{E}}_\cR$ is a compact subset of $\cT_\cR$ with non--empty interior, which  satisfies the conditions of Theorem~\ref{thm:gaussianWidth}. Also, since $\cT_\cR\cap \bA(\beta_{c_0})$ is a cone, the following can be easily verified:
\begin{equation}
\begin{split}
w_{\Omega,g}(\bar{\cE}_\cR\cap \bA(\beta_{c_0}))&=w_{\Omega,g}(\cE_\cR\cap \bA(\beta_{c_0}))\\
w_{G}(\bar{\cE}_\cR\cap \bA(\beta_{c_0}))&=w_{G}(\cE_\cR\cap \bA(\beta_{c_0}))\le w_{G}(\cE_\cR)
\end{split}
\end{equation}

We define a random variable $V(\Omega)=\sup_{X\in {\cE}_\cR\cap \bA(\beta_{c_0})}\Big|\frac{d_1d_2}{|\Omega|}\|P_\Omega(X)\|_2^2-1\Big|$.

Note that: 
for $X\in{\c{E}}_{\c{R}}\cap\b{A}(\beta_{c_0})$, $\bE\frac{d_1d_2}{|\Omega|}\|\c{P}_\Omega(X)\|^2=1$; and \\
for $X\in\bar{\c{E}}_{\c{R}}\cap\b{A}(\beta_{c_0})$, $\|X\|_\infty \le\frac{\beta_{c_0}}{\sqrt{d_2d_2}}\|X\|_F^2\le\frac{\beta_{c_0}}{\sqrt{d_2d_2}}$.

\subsubsection{Expectation of $V(\Omega)$}
Recall that $\Omega=\{E_\s:s=1,2,\ldots|\Omega|\}$ are sampled uniformly form standard basis for $\Rmn$, $(\epsilon_\s)$ are a sequence of independent Rademacher variables, and $w_G(.)$ denotes the Gaussian width. For constant $k_1,k_2,k_3$ not necessarily same in each occurrence:
\begin{flalign}
\nonumber &&\bE V(\Omega)&\overset{(a)}{\le}\frac{2d_1d_2}{|\Omega|}\bE\sup_{X\in \cE_\cR\cap \bA(\beta_{c_0})}\Big|\sum_{\s=1}^{|\Omega|}\innerprod{X}{E_\s}^2\epsilon_\s\Big|
\overset{(b)}\le
k_1\beta_{c_0}\frac{\sqrt{d_1d_2}}{|\Omega|}\,\bE\sup_{X\in {\cE}_\cR\cap \bA(\beta_{c_0})}\Big|\sum_{\s=1}^{|\Omega|} \innerprod{X}{E_\s}\epsilon_{\s}\Big|&\\
&&&=k_1\beta_{c_0}\frac{\sqrt{d_1d_2}}{|\Omega|}w_{\Omega,\epsilon}({\bar{\cE}}_\cR\cap \bA(\beta_{c_0}))\overset{(c)}{\le} k_1\sqrt{\frac{\beta_{c_0}^2w^2_G(\cE_\cR)}{|\Omega|}} +k_2\frac{\beta^2_{c_0}}{{|\Omega|}}\overset{(d)}{\le} \frac{k_3}{c_0\sqrt{\log d}},&\label{eq:Ev1}
\end{flalign}
where $(a)$ follows from  symmetrization (Lemma \ref{lem:symmetrization}), $(b)$ from contraction principle as $\phi_{k}(\innerprod{X}{E_\s})=\frac{\innerprod{X}{E_\s}^2}{2\sup_{X\in\cE_\cR\cap\bA(\beta_{c_0})}\|X\|_\infty}$ is a contraction~(Lemma\ref{lem:contraction}),  $(c)$ follows from Theorem~\ref{thm:gaussianWidth}, and $(d)$ using $|\Omega|>c_0^2w_G^2(\c{E}_\c{R})\log{d}$.

\subsubsection{Concentration about $\bE V(\Omega)$}
Given $\Omega$, let $\Omega^\prime\subset [m]\times [n]$ be another set of indices that differ from $\Omega$ in exactly one element. We have:
\begin{align}
\nonumber V(\Omega)-V(\Omega^\prime)&=\sup_{X\in\cE_\cR\cap \bA(\beta_{c_0})}\Big|\frac{d_1d_2}{|\Omega|}\sum_{ij\in\Omega}X_{ij}^2-1\Big|-\sup_{X\in\cE_\cR\cap \bA(\beta_{c_0})}\Big|\frac{d_1d_2}{|\Omega|}\sum_{kl\in\Omega^\prime}X_{kl}^2-1\Big|\\
\nonumber &\le\frac{d_1d_2}{|\Omega|}\sup_{X\in\cE_\cR\cap \bA(\beta_{c_0})}\left(\Big|\sum_{ij\in\Omega}X_{ij}^2-\sum_{kl\in\Omega^\prime}X_{kl}^2\Big|\right)\\&\le\frac{2d_1d_2}{|\Omega|}\sup_{X\in\cE_\cR\cap \bA(\beta_{c_0})}\|X\|_\infty^2\le\frac{2\beta_{c_0}^2}{|\Omega|}.
\end{align}
By similar arguments on $V(\Omega^\prime)-V(\Omega)$, $|V(\Omega)-V(\Omega^\prime)|\le \frac{2\beta_{c_0}^2}{|\Omega|}$.
Therefore, using Mc Diarmid's inequality \eqref{eq:mcdiarmid}, we have $P(V(\Omega)>\mathbb{E}V(\Omega)+\delta)\le \exp\left(-c_1^\prime \frac{\delta^2|\Omega|}{\beta_{c_0}^4}\right)$. 
Using $\delta=\frac{1}{c_0\sqrt{\log d}}$, we have
\[P\Bigg(V(\Omega)>\frac{k_3^\prime }{{c}_0\sqrt{\log d}}\;\Bigg)\le \exp{\Big(-c_1w_G^2(\cE_\cR)\Big)},\]

where $c_0$ is a constant that can be chosen independent of $k_3$. Choosing $c_0$ large enough, we can set $\kappa_{c_0}:=1-\delta_{c_0}=1-\frac{k_3^\prime}{c_0\sqrt{\log{d}}}$ close to $1$.
\mybox
\subsection{Proof of Lemma~\ref{lem:lambdacn}}
Recall that $\w\in\bR^{|\Omega|}$ is a vector of centered, unit variance sub-Gaussian random variables with $\|\w_\s\|_{\Psi_2}\le b$. Combining Lemma~\ref{lem:subgaussSquare} and Lemma~\ref{lem:centering}, we have that $\w_\s^2$ and $\w_\s^2-1$ are sub--exponential with $\|\w_\s^2-1\|_{\Psi_1}\le 2\|\w_\s^2\|_{\Psi_1}\le4\|\w_\s\|_{\Psi_2}\le4b^2$. Thus, using Lemma~\ref{lem:subExpBernstein}, for a constant $c_2^\prime$, we have:
\begin{equation}
\bP\Big(\Big|\frac{1}{|\Omega|}\sum_{\s=1}^{|\Omega|}\w_\s^2-1\Big|>\tau\Big)\le 2\exp{\Big(-c_2^\prime|\Omega|\min{\big\{\frac{\tau^2}{16b^4},\frac{\tau}{4b^2}\big\}}\Big)}.
\end{equation}
Choosing $\tau$ to be an appropriate constant, we have $\|P_\Omega(\Theta^*)-y)\|_2\le 2\nu\sqrt{|\Omega|}\le \lambda_\cn$ w.p. greater than $1-\exp(-c_2\tau|\Omega|)$, and the lemma follows from the optimality of $\hat{\Theta}_\cn$ and triangle inequality.

\section{Appendix to Proof of Theorem \ref{thm:gaussianWidth}} \label{app:thm2}
\subsection{Results from Generic Chaining}

 In this section, $K$ denotes a universal constant, not necessarily the same at each occurrence.  
\begin{mydefinition}[Gamma Functional (Definition $2.2.19$ in \citep{talagrand2014upper})] \normalfont
Given a complete pseudometric space $(T,d)$, an \textit{admissible sequence} is an increasing sequence $(\c{A}_n)$ of partitions of $T$ such that $|\c{A}_0|=1$ and $|\c{A}_n|\le 2^{2^n}$ for $n\ge1$. For $\alpha>0$, we define the Gamma functional $\gamma_{\alpha}(T,d)$  as follows: 
\begin{equation}
\gamma_\alpha(T,d)=\inf_{(\c{A}_n)_{n\ge 0}}\sup_{t\in T}\sum_{n\ge0}2^{n/\alpha}\Delta_d(A_n(t)), 
\label{eq:gamma2}
\end{equation}
where $\inf$ is over all admissible sequences $(\c{A}_n)$, $A_n(t)$ is the unique element of $\c{A}_n$ that contains $t$, and $\Delta_d(A)$ is the diameter of the set $A$ measured in metric $d$.
\end{mydefinition}

\begin{lemma} [Majorizing Measures Theorem (Theorem $2.4.1$ in \citet{talagrand2014upper})] \label{lem:mm} Given a closed set $T$ in a metric space, let $(X_t)_{t\in T}$ be a centered Gaussian process  indexed by $t\in T$, i.e. $(X_t)$ are jointly Gaussian. For $s,t\in T$, let $d_X(s,t):=\sqrt{\bE(X_s-X_t)^2}$ denote the canonical pseudometric associated with $(X_t)$. We then have : 
\[ \frac{1}{K}\gamma_2(T,d_X)\le \bE\sup_{t\in T} X_t\le K\gamma_2(T,d_X).
\]
In particular, considering the canonical Gaussian process $(\sum_it_ig_i)_{t\in T}$, we have:
\[ \frac{1}{K}\gamma_2(T,\|.\|_F)\le w_G(T)\le K\gamma_2(T,\|.\|_F).
\]
\end{lemma}

\begin{lemma}[Theorem $2.4.12$ in \citet{talagrand2014upper}] \label{lem:subGaussianGaussian} Let $(X_t)_{t\in T}$ be a centered Gaussian process with canonical distance $d_X=\sqrt{\bE(X_s-X_t)^2}$. Let $(Y_t)_{t\in T}$ be another centered process indexed by the same set $T$, such that it satisfies the following condition: 
\[\forall s,t\in T, u>0, \quad\bP(|Y_s-Y_t|>u)\le 2\exp{\Big(-\frac{u^2}{2d^2_X(s,t)}\Big)},\]
then, we have $\bE\sup_{s,t\in T}|Y_s-Y_t|\le K\bE\sup_{t\in T} X_t$. 

If further, $(Y_t)_{t\in T}$ is symmetric, then 
$\bE\sup_t |Y_t|\le \bE\sup_{s,t\in T}|Y_s-Y_t|=2\bE\sup_{t\in T}Y_t$.

\end{lemma}
\noindent \textbf{Note:} From the definition of sub--Gaussian random variables (Section \ref{sec:subGauss}), using the above lemma,  sub--Gaussian complexity measures can be directly bounded by Gaussian complexities.

\begin{lemma} [Theorem $3.1.4$ in \citet{talagrand2014upper}] \label{lem:dw}Let $T$ be a compact group with non--empty interior. Consider a translation invariant random distance $d_\omega$, that depends on a random parameter $\omega$ and let 
${d(s,t)=\sqrt{\bE d_\omega^2(s,t)}}$, then :
\[\big(\bE \gamma_2^2(T,d_\omega)\big)^{1/2}\le K\gamma_2(T,d)+K\big(\bE \sup_{s,t\in T}d_\omega^2(s,t)\big)^{1/2}\]
\end{lemma}



\subsection{Proof of Lemma \ref{lem:gammaResult}}\label{app:gammaResult}
\textbf{Statement of Lemma~\ref{lem:gammaResult}} \\For a compact subset $S\subseteq\Rmn$ with non--empty interior, $\exists$ constants $k_1$, $k_2$ such that: \\
{$\displaystyle
\nonumber w_{\Omega,g}({S})=\bE\sup_{X\in {S}}\mathcal{X}_{\Omega,g}(X)\le  k_1\sqrt{\frac{|\Omega|}{d_1d_2}}w_G(S)+k_2\sqrt{\bE\sup_{X,Y\in S}\|P_\Omega(X-Y)\|_2^2}.
$\mybox

\textbf{Proof: } Recall definition of $(\mathcal{X}_{\Omega,g}(X))_{X\in S}$ from \eqref{eq:rProc}, such that $\mathcal{X}_{\Omega,g}(X)=\sum_{\s} \langle X,E_\s\rangle g_\s$.

By Fubini's theorem $\bE_{\Omega,g} \sup_{X\in S}\mathcal{X}_{\Omega,g}(X)=\bE_\Omega\bE_g\sup_{X\in S}\mathcal{X}_{\Omega,g}(X)$. 

Also, we have the following results:
\begin{compactitem}
\item Given a random variable $\Omega$, $(\mathcal{X}_{\Omega,g}(X))$ is a Gaussian process  with a translation invariant canonical distance given by $d_\Omega(X,Y)=\|P_\Omega(X-Y)\|_2^2$. 
\item $d(X,Y):=\sqrt{\bE_\Omega d_\Omega^2(X,Y)}=\sqrt{\p}\|X-Y\|_F$
\end{compactitem}

Using Lemma~\ref{lem:mm} we have: $\bE_g\sup_{X\in S}\mathcal{X}_{\Omega,g}(X)\le K\gamma_2(S,d_\Omega)$, and the following holds:
\begin{flalign}
\nonumber w_{\Omega,g}(S)&=\bE_\Omega\bE_g\sup_{X\in S}\mathcal{X}_{\Omega,g}(X)\le K \bE_\Omega \gamma_2(S,d_\Omega)\overset{(a)}{\le}\sqrt{\bE_\Omega \gamma_2^2(S,d_\Omega)}&\\
&\overset{(b)}{\le} K\sqrt{\p}\gamma_2(S,\|.\|_F)+K\sqrt{\bE\sup_{X,Y\in S}\|P_\Omega(X-Y)\|_2^2},&\label{eq:bt1}
\end{flalign}
where $(a)$ follows from Jensen's inequality, $(b)$ from Lemma~\ref{lem:dw} and noting that from definition $\forall M>0$, $\gamma_2(T,M\dot d)=M\gamma_2(T,d)$. Lemma~\ref{lem:gammaResult} now follows from \eqref{eq:bt1} and Lemma~\ref{lem:mm}. \mybox 

\subsection{Proof of Lemma~\ref{prop:temp}}
\textbf{Statement of Lemma~\ref{prop:temp}}\\
There exists constants $k_3$, $k_4$, such that for compact $S\subseteq\Bmn$ with non--empty interior
$$\bE\sup_{X,Y\in S}\|P_\Omega(X-Y)\|_2^2\le k_3\p w_G^2(S)+k_4\sup_{X,Y\in S}\|X-Y\|_\infty w_{\Omega,g}(S)$$

\textbf{Proof: } Using triangle inequality, we have: 
\begin{equation}
\bE\sup_{X,Y\in S}\|P_\Omega(X-Y)\|_2^2\le \bE\sup_{X,Y\in S}|\|P_\Omega(X-Y)\|_2^2-\bE\|P_\Omega(X-Y)\|_2^2| + \sup_{X,Y\in S}\bE\|P_\Omega(X-Y)\|_2^2
\label{eq:bl91}
\end{equation}
Further,  
\begin{equation}
\sup_{X,Y\in S}\bE\|P_\Omega(X-Y)\|_2^2=\p \sup_{X,Y\in S}\|X-Y\|_F^2\le \p \gamma_2^2(S,\|.\|_F),
\label{eq:bl92}
\end{equation}
where the last inequality follows from the definition of $\gamma_\alpha$.

Finally, we have the following set of equations:
\begin{flalign}
\nonumber &\bE\sup_{X,Y\in S}\big|\|P_\Omega(X-Y)\|_2^2-\bE[\|P_\Omega(X-Y)\|_2^2]\big|
= \bE\sup_{X,Y\in S}\big|\sum_{\s=1}^{|\Omega|}\innerprod{X-Y}{E_\s}^2-\bE{\innerprod{X-Y}{E_\s}}^2\big|&\\
\nonumber &\overset{(a)}{\le}2\bE_{\Omega,(\epsilon_s)}\sup_{X,Y\in S}|\sum_{\s=1}^{|\Omega|}\innerprod{X-Y}{E_\s}^2\epsilon_\s|
\overset{(b)}{\le}k_4^\prime\sup_{X\in S}\|X-Y\|_\infty\bE_{\Omega,g}\sup_{X,Y\in S}|\sum_{\s=1}^{|\Omega|}\innerprod{X-Y}{E_\s}g_\s|&\\
&\overset{(c)}{\le} 2k_4^\prime\sup_{X,Y\in S}\|X-Y\|_\infty\bE_{\Omega,g}\sup_{X\in S}|\sum_{\s=1}^{|\Omega|}\innerprod{X}{E_\s}g_\s| \overset{(d)}{\le} 4k_4^\prime\sup_{X,Y\in S}\|X-Y\|_\infty w_{\Omega,g}(S),
\label{eq:linfty}
\end{flalign}
where $(\epsilon_\s)$ are standard Rademacher variables, i.e. $\epsilon_\s\in\{-1,1\}$ with equal probability, $(a)$ follows from  symmetrization argument (Lemma \ref{lem:symmetrization}), $(b)$ follows from contraction principles  Lemma~\ref{lem:contraction} and using $\phi(\innerprod{X}{E_\s})=\frac{\innerprod{X}{E_\s}^2}{2\sup_{X\in S}\|X\|_\infty}$ as a contraction, $(c)$ follows from triangle inequality, and $(d)$ follows from $g_\s$ being symmetric (Lemma $2.2.1$ in \citet{talagrand2014upper}). \mybox

The lemma follows by combining Lemma~\ref{lem:mm} and equations \eqref{eq:bl91}, \eqref{eq:bl92}, and \eqref{eq:linfty}. 

\subsection{Remaining Steps in the Proof of Theorem~\ref{thm:gaussianWidth}}\label{app:remainingsteps}From Lemma \ref{prop:temp}, we have the following:
\begin{flalign}
\nonumber \sqrt{\bE\sup_{X,Y\in S}\|P_\Omega(X-Y)\|_2^2}&\overset{(a)}{\le} K_3\sqrt{\p} w_G(S)+\sqrt{k_4(\sup_{X,Y\in S}\|X-Y\|_\infty )w_{\Omega,g}(S)}&\\
 &\overset{(b)}\le K_3\sqrt{\p} w_G(S)+K_4(\sup_{X,Y\in S}\|X-Y\|_\infty )+\frac{1}{2}w_{\Omega,g}(S),&
\label{eq:linfty2}
\end{flalign}
where $(a)$ follows from triangle inequality, $(b)$ using $\sqrt{ab}\le a/2+b/2$. 

Bound on $w_{\Omega,g}(S)$ in Theorem~\ref{thm:gaussianWidth} follows by using \eqref{eq:linfty2} in Lemma~\ref{lem:gammaResult}.

\section{Spectral $\mathbf{k}$--Support Norm}
Recall the following definition of spectral $k$--support norm  $\|\Theta\|_\ksp$ from \eqref{eq:ksupp}:
\begin{equation}
\|\Theta\|_\ksp=\inf_{v\in\cV(\mathcal{G}_k)}\Big\{\sum_{g\in\cG_k}\|v_g\|_2:\sum_{g\in\cG_k}v_g=\sigma(\Theta)\Big\},
\end{equation}
where $\cG_k=\{g\subseteq [\bar{d}]: |g|\le k\}$ is  the set of all subsets $[\bar{d}]$ of cardinality at most $k$, and  $\cV(\mathcal{G}_k)=\{(v_g)_{g\in\cG_k}: v_g\in\bR^{d_1}, \text{supp}(v_g)\subseteq g\}$. 
\begin{proposition}[Proposition $2.1$ in \citet{argyriou2012sparse}] For $\Theta\in\bR^{\bar{d}\times \bar{d}}$ with singular values  $\sigma(\Theta)=\{\sigma_1,\sigma_2,\ldots,\sigma_{\bar{d}}\}$, such that $\sigma_1\ge\sigma_2\ge\ldots,\ge\sigma_{\bar{d}}$. Then,
\begin{equation}
\|\Theta\|_\ksp=\Bigg(\sum_{i=1}^{k-r-1}\sigma_i^2+\frac{1}{r+1}\Bigg(\sum_{i=k-r}^{\bar{d}}\sigma_i\Bigg)^2\Bigg)^\frac{1}{2},
\label{eq:ksupp-alternate}
\end{equation}
where $r\!\in\!\{0,1,2,\ldots,k\!-\!1\}$ is the unique integer satisfying
$\sigma_{k-r-1}>\frac{1}{r+1}\sum_{i=k-r}^{d_1}\sigma_i\ge\sigma_{k-r}.
$\mybox
\end{proposition}
\subsection{Proof of Lemma~\ref{lem:ksupp1}}
\textbf{Statement of Lemma~\ref{lem:ksupp1}}\\
If rank of $\Theta^*$ is $s$ and
$\cE_\cR$ is the error set from $\cR(\Theta)=\|\Theta\|_\ksp$, then
\[w_G^2(\cE_\cR)\le s(2\bar{d}-s)+\Big(\frac{(r+1)^2\|\sigma^*_{I_2}\|_2^2}{\|\sigma^*_{I_1}\|_1^2}+|I_1|\Big)(2\bar{d}-s).\]
\mybox

\textbf{Proof } We state the following lemmas from existing work.
\begin{lemma}[Equation $\boldsymbol{60}$ in \citet{richard2014tight}]Let $z$ be an $s\ge k$ sparse vector in $\bR^p$, and let $\tilde{z}$ is the vector $z$ sorted in non increasing order of $|z_i|$. Denote $r\in\{0,1,2,\ldots,k-1\}$ to be the unique integer satisfying
\[|\tilde{z}_{k-r-1}|>\frac{1}{r+1}\sum_{i=k-r}^{p}|\tilde{z}_i|\ge|\tilde{z}_{k-r}|.\]
Define $I_2=\{1,2,\ldots,k-r-1\}$, $I_1=\{k-r,k-r+1,\ldots,s\}$, and $I_0=\{s+1,s+2,\ldots,p\}$; and let $\tilde{z}_I$ denote the vector $\tilde{z}$ restricted to indices in $I$. Then the sub--differential of the  vector $k$--support norm denoted by $\|.\|_{\text{vk-sp}}$ at $w$ is given by:
\[\partial\|z\|_{\text{vk-sp}}=\frac{1}{\|z\|_{\text{vk-sp}}}\Big\{\tilde{z}_{I_2}+\frac{1}{r+1}\|\tilde{z}_{I_1}\|_1(\text{sign}(\tilde{z}_{I_1})+h_{I_0}): \|h\|_\infty\le1\Big\},\]
\label{lem:subdiff1}
\end{lemma}
\begin{lemma} [Theorem $2$ in \citet{watson1992characterization}]
Let $\cR:\Rmn\to\bR_+$ be an orthogonally invariant norm; i.e.~$\cR(X)=\phi(\sigma(X))$ such that $\phi:\bR^{d_1}\to \bR_+$ is a symmetric gauge function satisfying: 
\begin{inparaenum}[(a)]
\item $\phi(x)>0\;\forall x\ne0$, \item $\phi(\alpha x)=|\alpha|\phi(x)$, \item $\phi(x+y)\le\phi(x)+\phi(y)$, and \item $\phi(x)=\phi(|x|)$.
\end{inparaenum} 

Further let $\partial\phi(x)$ denote the sub--differential of $\phi$ at $x$. Then for $X\in\bR^{\bar{d}\times \bar{d}}$ with singular value decomposition (SVD) $X=U_X\Sigma_XV_X^\top$ and $\sigma_X=\text{diag}(\Sigma_X)$, the sub--differential of $\cR(X)$ is given by:
\[\partial\cR(X)=\{U_XDV_X^\top: D=\text{diag}(d),\text{ and }d\in\partial\Phi(\sigma_X)\}.\]
\label{lem:subdiff}
\end{lemma}

Since spectral $k$--support norm of a matrix $X=U_X\Sigma_XV_X^\top$ is the vector $k$--support norm applied to the singular values $\sigma_X=\text{diag}(\Sigma_X)$, Lemma~\ref{lem:subdiff1} and \ref{lem:subdiff}  can be used to infer the following:
\begin{equation}
\partial\|X\|_\ksp\!=\!\Big\{U_XDV_X^\top:\text{diag}(D)\in\frac{1}{\|\sigma_X\|_{\text{vk-sp}}}\Big\{\sigma_{X_{I_2}}+\frac{\|\sigma_{X_{I_1}}\|_1}{r+1}(\boldsymbol{1}_{I_1}+h_{I_0}): \|h\|_\infty\le1\Big\}\Big\}.
\label{eq:subdiff2}
\end{equation}
where $\boldsymbol{1}\in\bR^{\bar{d}}$ denotes a vector of all ones.

\noindent The error cone for $\cR(.)=\|.\|_\ksp$ is given by the tangent cone:
\[\cT_\cR=\text{cone}\{\Delta:\|\Theta^*+\Delta\|_\ksp\le\|\Theta^*\|_\ksp\},\] and the polar of the tangent cone -- the \textit{normal cone} is given by 
\[\cT_\cR^*=\cN_\cR(\Theta^*)=\{Y:\innerprod{Y}{X}\le 0\;\forall X\in\cT_\cR\}=\text{cone}(\partial\cR(\Theta^*))\]

\noindent Let $\Theta^*=U^*\Sigma^*V^{*\top}$ be the full SVD of $\Theta^*$, such that $\sigma^*=\text{diag}(\Sigma^*)\in \bR^{\bar{d}}$ and $\sigma^*_1\ge\sigma_2^*\ldots\ge\sigma^*_{\bar{d}}$. Let $u^*_i$ and $v^*_i$ for $i\in[\bar{d}]$ denote the $i^\text{th}$ column of $U^*$ and $V^*$, respectively.
Further, let the rank of $\Theta^*$ be $\text{rk}(\Theta^*)=\|\sigma^*\|_0=s$. 

Like for the vector case, denote $r\in\{0,1,2,\ldots,k-1\}$ to be the unique integer satisfying
$\displaystyle \sigma^*_{k-r-1}>\frac{1}{r+1}\sum_{i=k-r}^{p}\sigma^*_i\ge\sigma^*_{k-r}.$
Define $I_2=\{1,2,\ldots,k-r-1\}$, $I_1=\{k-r,k-r+1,\ldots,s\}$, and $I_0=\{s+1,s+2,\ldots,p\}$; 
Also define the subspace:
\[T=\text{span}\{u^*_ix^\top:i\in I_2\cup I_1, x\in\bR^{\bar{d}}\}\cup\text{span}\{yv_i^{*\top}:i\in I_2\cup I_1, y\in\bR^{\bar{d}}\}
\]
Let $T^\perp$ be the subspace orthogonal to $T$ and let $P_T$ and $P_{T^\perp}$ be the projection operators onto $T$ and $T^\top\!\!\!\!,\;$ respectively. From \eqref{eq:subdiff2} we have,
\[\cN_\cR(\Theta^*)=\Bigg\{Y=U^*DV^{*\top}: D=\text{diag}\Big(t\frac{r+1}{\|\sigma_{I_1}^*\|_1}\sigma^*_{I_2}+t\boldsymbol{1}_{I_1}+th_{I_0}\Big): t\ge 0, \|h\|_\infty\le1\Bigg\},\]

\noindent Finally, from Lemma~\ref{lem:cone}, we have that 
\begin{flalign*} &&&w_G^2(\cT_\cR\cap\bS^{\bar{d}\bar{d}-1})\le\bE_G\inf_{X\in\cN_\cR(\Theta^*)}\|G-X\|_F^2&\\
&&&\le\bE_G\inf_{\substack {t>0\\ \|h\|_\infty\le1}}\Big\|P_T(G)-t\frac{r+1}{\|\sigma_{I_1}^*\|_1}\sum_{i\in I_2}\sigma^*_iu^*_iv^{*\top}_i+t\sum_{i\in I_1}u_i^*v_i^{*\top}+P_{T^\perp}(G)-t\sum_{i\in I_0}h_iu^*_iv_i^{*\top}\Big\|_F^2&
\end{flalign*}
Let $P_{T^\perp}(G)=\sum_{i\in I_0}\sigma_i(P_{T^\perp}G)u_i^*v_i^{*\top}$ be the decomposition of $P_T^\perp(G)$ in the basis of of $\{u_i^*v_i^{*\top}\}_{i\in I_0}$. Taking $t=\|P_{T^\perp}(G)\|_\op=\max_{i\in I_0}\sigma_i(P_{T^\perp}(G))$, and $h_i=\sigma_i(P_{T^\perp}(G))/\|P_{T^\perp}(G)\|_\op\le 1$, we have:
\begin{equation}
w_G^2(\cT_\cR\cap\bS^{\bar{d}\bar{d}-1})\le\bE_G\|P_T(G)\|_F^2+\Bigg(\frac{(r+1)^2\|\sigma^*_{I_2}\|_2^2}{\|\sigma^*_{I_1}\|_1^2}+|I_1|\Bigg)\bE_G\|P_T(G)\|_2^2.
\end{equation}
Lemma~\ref{lem:ksupp1} follows by using $\bE_G\|P_T(G)\|_F^2=s(2\bar{d}-s)$ and $\bE_G\|P_T(G)\|_\op^2\le 2(2\bar{d}-s)$ from \citet{chandrasekaran2012convex}.



\section{Preliminaries}\label{sec:prelims}
\subsection{Probability and Concentration}
\begin{lemma}[Bernstein's Inequality (moment version)] Let $X_i, i=1,2,\ldots,N$ be independent zero mean random variables. Further, let $\sigma^2=\sum_i \bE[X_i^2]$, and $M>0$ be such that the following moment conditions are satisfied for $p\ge2$,
\[ \bE[X_i^p]\le \frac{p!\sigma^2 M^{p-2}}{2}
\]
Then the following concentration inequality holds:
\begin{equation}
\bP\Big(\Big|\sum_iX_i\Big|>u\Big)\le 2\exp{\Big(\frac{-u^2}{2\sigma^2+2Mu}\Big)}
\label{eq:Bernstein1}
\end{equation}
\end{lemma}

\begin{lemma}[McDiarmid's Inequality] Let $X_i, i=1,2,\ldots,N$ be independent random variables. Consider a function $f:\mathbb{R}^N\to \bR$:
\begin{flalign}
\nonumber &&\text{If}\quad\quad&\forall i,\;\sup_{X_1,X_2,\ldots,X_N,X_i^\prime} |f(X_1,X_2,\ldots,X_N)-f(X_1,X_2,\ldots,X_{i-1},X_i^\prime,X_{i+1},\ldots,X_N)\le c_i,&\\
&&\text{then, }\;\; &\bP(|f(X_1,X_2,\ldots,X_N)-\bE f(X_1,X_2,\ldots,X_N)|>u)\le2\exp\Big(\frac{-2u^2}{\sum_i c_i^2}\Big)&
\label{eq:mcdiarmid}
\end{flalign}
\end{lemma}

\begin{lemma}[Symmetrization (Lemma $6.3$ in \citet{ledoux1991probability})] Let $F:\bR_+\to\bR_+$ be a convex function, and $X_i, i=1,2,\ldots $ be a sequence of mean zero random variables in a Banach space $B$, s.t $\forall i, \bE F\|X_i\|<\infty$. Denote a vector of standard Rademacher variables of appropriate dimension  as $(\epsilon_i)$, then 
\begin{equation}
\bE F\Big(\frac{1}{2}\|\sum_i\epsilon_iX_i\|\Big)\le \bE F\|\sum_iX_i\|\le \bE F\Big(2\|\sum_i\epsilon_iX_i\|\Big)
\end{equation}
Further, if $X_i$ are not centered, then $\bE F\Big(\|\sum_iX_i-\bE[X_i]\|\Big)\le \bE F\Big(2\|\sum_i\epsilon_iX_i\|\Big)$
\label{lem:symmetrization}
\end{lemma}

\begin{lemma}[Contraction Principle] Consider a bounded  $T\subset \bR^N$, a standard Gaussian and standard Rademacher sequence, $(g_i)\in\bR^N$ and  $(\epsilon_i)\in\bR^N$, respectively. If $\phi_i:\bR\to\bR$, $i\le N$ are contractions, i.e. $\forall s,t\in \bR$, $|\phi_i(s)-\phi_i(t)|\le|s-t|$, and  with $\phi_i(0)=0$, then for any convex function $F:\bR_+\to\bR_+$, the following results are from  Corollary $3.17$, Theorem $4.12$, and Lemma $4.5$, respectively in \citet{ledoux1991probability}:
\begin{flalign}
&\bE F\Big(\frac{1}{2}\sup_{t\in T}\Big|\sum_{i=1}^Ng_i\phi_i(t_i)\Big|\Big)\le \bE F\Big(2\sup_{t\in T}\Big|\sum_{i=1}^Ng_it_i\Big|
\Big)&\\
&\bE F\Big(\frac{1}{2}\sup_{t\in T}\Big|\sum_{i=1}^N\epsilon_i\phi_i(t_i)\Big|\Big)\le \bE F\Big(2\sup_{t\in T}\Big|\sum_{i=1}^N\epsilon_it_i\Big|
\Big)&\\
&\bE F\Big(\|\sum_{i=1}^N\epsilon_it_i\|\Big)\le \bE F\Big(\sqrt{\frac{\pi}{2}}\|\sum_{i=1}^Ng_it_i\|\Big)
\end{flalign}
\label{lem:contraction}
\end{lemma}

\subsection{Gaussian Width}\label{sec:gwidth} 
Gaussian width plays a key role high dimensional estimation, and plenty of tools have been developed for computing Gaussian widths of compact subsets \citep{dudley1967sizes,ledoux1991probability,talagrand2014upper, chandrasekaran2012convex}. The existing work is specially well adapted for computing Gaussian widths for intersection of convex cones with unit norm balls \citep{chandrasekaran2012convex}, and  recent work of \citet{banerjee2014estimation} propose a mechanism for exploiting these tools for arbitrary compact sets.  We briefly note some of the key results that aid in  computing Gaussian widths. Recall that $\Smn$ is a unit Euclidean sphere in $\Rmn$. Further, for a cone $\mathcal{C}\in\Rmn$, we define the polar cone as $\mathcal{C}^\circ=\{X:\innerprod{X}{Y}\le0,\forall Y\in\mathcal{C}\}$. 

\subsubsection{Direct Estimation}
The Gaussian width of a compact set $T$ can be directly estimated as a supremum of Gaussian process over dense countable subset $\bar{T}$ of $T$ as
$w_G(T)=\sup_{X\in\bar{T}} \innerprod{X}{G}.$

We state the following properties are often used in direct estimation. These properties are consolidated from  \citet{talagrand2014upper}, \citet{chandrasekaran2012convex} and \citet{banerjee2014estimation}.  In the following statements, $k$ is a constant not necessarily the same in each occurrence:
\ibullet{
\item Translation invariant and homogeneous: for any $a\in\bR$, $w_G(S+a)=w_G(S)$; and . 
\item $w_G(\text{conv}(T))\le w_G(T)$
\item $w_G(T_1+T_2)\le w_G(T_1)+w_G(T_2)$
\item If $T_1\subseteq T_2$, then $w_G(T_1)\le w_G(T_2)$.
\item If $T_1$ and $T_2$ are convex, then $w_G(T_1\cup T_2)+w_G(T_1\cap T_2)=w_G(T_1)+w_G(T_2)$}

\subsubsection{Dudley's Inequality and Sudakov Minorization}
\begin{mydefinition}[Covering Number] \normalfont Consider a metric $d$  defined on  $S\subset\Rmn$. Given $\epsilon\!>\!0$, the $\epsilon$--covering number of $S$ with respect to $d$, denoted by $\cN(S,\epsilon,d)$, is the minimum number of points $\{\bar{X}_1,\bar{X}_2,\ldots,\bar{X}_{\cN(S,\epsilon,d)}\}$ such that $\forall$ $X\in S$, there exists $i\in\{1,2,\ldots,\cN(S,\epsilon,d)\}$ with $d(X,\bar{X}_i)\le\epsilon$. The set $\{\bar{X}_1,\bar{X}_2,\ldots,\bar{X}_{\cN(S,\epsilon,d)}\}$ is called the $\epsilon$--cover of $S$.
\end{mydefinition}

\begin{lemma}[Dudley's Inequality and Sudakov Minoration] If $S$ is compact, then for any $\epsilon>0$, there exists a constant $c$ s. t.
\[c\epsilon\sqrt{\log{N(S,\epsilon,\|.\|_F)}}\le w_G(S)\le24\int_0^\infty \sqrt{N(S,\epsilon,\|.\|_F)}\text{d}\epsilon.\]
The upper bound is the Dudley's inequality and lower bound is by Sudakov minoration.
\end{lemma}

\subsubsection{Geometry of Polar Cone}
\begin{lemma}[Proposition $\boldsymbol{3.6}$ and Theorem $\boldsymbol{3.9}$ of \citet{chandrasekaran2012convex}] \label{lem:cone} If $\mathcal{C}\subset \Rmn$ is a non--empty convex cone and $\mathcal{C}^\circ$ be its  polar cone, then:\normalfont 
\enum{\item [Distance to polar cone]: 
$\displaystyle{w_G(\mathcal{C}\cap\Smn)\le \bE_G[\inf_{X\in\mathcal{C}^\circ}\|G-X\|_F]}$
\item [Volume of polar cone]: $\displaystyle{w_G(\mathcal{C}\cap\Smn)\le 3\sqrt{\frac{4}{\text{vol}(C^\circ\cap\Smn)}}}$}
\end{lemma}

\subsubsection{Infimum over Translated Cones} 
\begin{lemma}[Lemma $\boldsymbol{3}$ of \citet{banerjee2014estimation}] Let $S\subset\Rmn$, and given $X\in S$, define $\rho(X)=\sup_{Y\in S}\|X-Y\|_F$ as the diameter of $S$ measured along $X$. Also define $\mathcal{G}(X)=\text{cone}(S-X)\cap\rho(X)\Bmn$, where $\Bmn$ is the unit Euclidean ball. Then,
\[w_G(S)\le\inf_{X\in S}w_G(\mathcal{G}(X))\]
\end{lemma}
\subsubsection{Generic Chaining}
Lemma~\ref{lem:mm} (from \citet{talagrand2014upper}) gives the tightest bounds on the Gaussian width of a set. The definition of $\gamma_2$  \eqref{eq:gamma2} can be used derive tight bounds on the Gaussian width that are optimal upto constants. Further results and  examples on using $\gamma$--functionals for Gaussian width computation can be found in the works of Talagrand \citep{talagrand1996majorizing,talagrand2001majorizing,talagrand2014upper}. 

\subsection{Sub--Gaussian and Sub--Exponential Random Variables}\label{sec:subGauss}
Recall the definition of sub--Gaussian random variables from Definition~\ref{def:subGauss}. 
\begin{mydefinition}[Sub--Exponential Random Variables]\label{def:subexp}
\normalfont A random variable $X$ is said be \textit{sub-exponential} if it satisfies one of the following equivalent conditions for $k_1$, $k_2$, and $k_3$ differing from one other by constants [Definition $5.13$ of \citet{vershynin2010introduction}].
\enum{
\item $\bP(|X|>t)\le e^{1-{{t}/{k_1}}}$, $\forall$ $t>0$, 
\item $\forall p\ge 1$,  $(\bE[|X|^p])^{1/p}\le k_2{p}$, or 
\item $\bE[e^{{X}/{k_3}}]\le e$.
}
The \textit{sub--exponential norm} is given by:
\begin{equation}
\|X\|_{\Psi_1}=\inf\Big\{t>0: \bE\exp\Big(\frac{|X|}{t}\Big)\le 2\Big\}=\sup_{p\ge1} p^{-1}(\bE[|X|^p])^{1/p}.
\end{equation}
\end{mydefinition}

\noindent The following results on sub--Gaussian and sub--exponential  variables are  from \citet{vershynin2010introduction}.
\begin{lemma}
[Hoeffding--type inequality, Proposition $\boldsymbol{5.10}$ in \citet{vershynin2010introduction}] \label{lem:subGaussHoeffding} Let $X_1,X_2,\ldots,X_N$ be independent centered sub-Gaussian random variables, and let $K = \max_i \|X_i\|_{\Psi_2}$. Then, ${\forall a\in\bR^N}$ and $t\ge0$, $\exists$ constant $c$ s.t., 
\begin{equation}
\bP\Big(\big|\sum_{i=1}^Na_iX_i\big|\ge t\Big)\le 2\exp{\Big(\frac{-ct^2}{K^2\|a\|_2^2}\Big)}.
\end{equation}
\end{lemma}

\begin{lemma}
[Bernstein--type inequality, Proposition $\boldsymbol{5.16}$ in \citet{vershynin2010introduction}] \label{lem:subExpBernstein}Let $X_1,X_2,\ldots,X_N$ be independent centered sub-exponential random variables, and let $K = \max_i \|X_i\|_{\Psi_1}$. Then $\forall a\in\bR^N$, and $t\ge0$, there exists a constant $c$ s.t.
\begin{equation}
\bP\Big(\big|\sum_{i=1}^Na_iX_i\big|\ge t\Big)\le 2\exp{\Big(-c\min{\Big\{\frac{t^2}{K^2\|a\|_2^2},\frac{t}{K\|a\|_\infty}\Big\}}\Big)}.
\end{equation}
\end{lemma}

\begin{lemma}[Lemma $\boldsymbol{5.14}$ in \citet{vershynin2010introduction}]\label{lem:subgaussSquare}
$X$ is sub--Gaussian if and only if $X^2$ is sub--exponential. Further, $\|X\|^2_{\Psi_2}\le\|X^2\|_{\Psi_1}\le2\|X\|^2_{\Psi_2}.$
\end{lemma}

\begin{lemma}[Remark $\boldsymbol{5.18}$ in \citet{vershynin2010introduction}]\label{lem:centering}
If $X$ is sub--Gaussian (or sub--exponential), then so is $X-\bE X$. Further, 
$\|X-\bE X\|_{\Psi_2}\le2\|X\|_{\Psi_2};\quad \|X-\bE X\|_{\Psi_1}\le2\|X\|_{\Psi_1}.$
\end{lemma}

\section{Extension to GLMs}\label{sec:glm}
This section provides directions for extending the work to matrix completion under generalized linear models. This section has not been rigorously formalized. An accurate version will be included in a longer version of the paper. 

We consider an observation model wherein the observation matrix $Y$ is drawn from a member of \textit{natural exponential family} parametrized by a structured ground truth matrix $\Theta^*$, such that:
\begin{flalign}
 P(Y|\Theta^*) &=\prod_{ij} p(Y_{ij})\, e^{Y_{ij}\Theta^*_{ij}-A(\Theta^*_{ij})},
\label{eq:expfam}
\end{flalign}
where $A:\text{dom}(\Theta_{ij})\to \bR$ is called the \textit{log--partition function} and is strictly convex and analytic, and $p(.)$ is called the \textit{base measure}. 
This family of distributions encompass a wide range of common distributions including Gaussian, Bernoulli, binomial, Poisson, and exponential among others.  
In a generalized linear matrix completion setting ~\citep{gunasekar2014exponential}, the task is to estimate $\Theta^*$ from a subset of entries $\Omega$ of $Y$, i.e. $(\Omega,P_\Omega(Y))$. 

A useful consequence of  exponential family distribution assumption for observation matrix  is that the negative log--likelihood loss over the observed entries is convex with respect to the natural parameter $\Theta^*$, and have a one-to-one correspondence with a rich class of divergence functions called the Bregman Divergence \citep{forster2002relative,banerjee2005clustering}.  
The negative log likelihood is proportional to: 
$${\cL_\Omega(\Theta)=\sum_{(i,j)\in\Omega} A(\Theta_{ij})-Y_{ij}\Theta_{ij}
}$$ 
We propose the following \textit{regularized matrix estimator} for generalized matrix completion:
\begin{equation}
\widehat{\Theta}_{\gds}=\argmin{{\|\Theta\|_\infty\le\frac{\alpha^*}{\sqrt{d_1d_2}}}}\frac{d_1d_2}{|\Omega|}\cL_\Omega(\Theta)+\lambda_{\gds}\cR(\Theta).
\label{eq:generalizedEstimatorDantzig}
\end{equation}
\begin{hypothesis} Let $\hat{\Theta}_\gds=\Theta^*+\hat{\Delta}_\gds$. In addition to the assumptions in Section~\ref{sec:setup}, we assume that for some $\eta\ge 0$,  $\nabla^2A(u)\ge e^{-\eta|u|}\forall\;u\in\mathbb{R}$. The following result holds for any fixed  $\gamma>1$. We define:
\begin{equation}
\widetilde\cT_{\cR,\gamma}=\text{cone}\{\Delta:\cR(\Theta^*+\Delta)\le \cR(\Theta^*)+\frac{1}{\gamma}\cR(\Theta^*)\},\quad\text{ and }\quad\widetilde\cE_{\cR,\gamma}=\widetilde\cT_{\cR,\gamma}\cap \Smn.
\label{eq:Tr1}
\end{equation}
Let $\lambda_\gds\ge\gamma\frac{{d_1d_2}}{|\Omega|}\cR^*(\nabla\cL_\Omega(\Theta^*))$, and for some $c_0$,  
$|\Omega|>\Big(\frac{\gamma+1}{\gamma-1}\Big)^{\!2}c_0^2w_G^2(\widetilde{\cE}_{\cR,\gamma})\log{d}$. There exists a constant $k_1$ such that for large enough $c_0$, there exists $\kappa_{c_0}>0$, such that with high probability, 
\[\|\widehat{\Delta}_\gds\|_F^2\!\le\! 4\alpha^{*2}\Big(\frac{\gamma+1}{\gamma-1}\Big)^{\!2}\!\max\!{\Bigg\{\!\frac{\lambda_\gds^2\Psi^2_\cR(\widetilde\cT_{\cR,\gamma})}{\zeta(\eta,\alpha^*)\kappa^2_{c_0}},\!\!\frac{c_0^2w_G^2(\widetilde{\cE}_{\cR,\gamma})\log{d}}{|\Omega|}\!\Bigg\}},\]
where $\zeta(\eta,\alpha^*)=e^{\frac{-4\eta\alpha^*}{\sqrt{d_1d_2}}}$, and  $\alpha^*$, $w_G(.)$, and $\Psi_\cR(.)$ are notations from Section~\ref{sec:main}.

\normalfont The conjectures follows by combining the results in this paper along with the results from \citet{banerjee2014estimation}, and \citet{gunasekar2014exponential}. This result is beyond the scope of this paper and will be dealt with more rigorously in a longer version of the paper. 

\end{hypothesis}

\end{document}